%% file: main.tex
\pdfoutput=1

\documentclass[11pt]{article}

\usepackage[preprint]{acl}

\usepackage{times}
\usepackage{latexsym}

\usepackage[T1]{fontenc}

\usepackage[utf8]{inputenc}

\usepackage{microtype}

\usepackage{inconsolata}

\usepackage{enumitem}
\usepackage{graphicx}
\usepackage{booktabs}
\usepackage{forest, paralist}
\usepackage{tikz}
\usetikzlibrary{shapes.geometric, arrows, positioning, fit, shadows, backgrounds, calc, decorations.pathreplacing, matrix}

\usepackage{ulem}
\usepackage{soul}

\definecolor{headerColor}{RGB}{74, 111, 165}
\definecolor{tableColor}{RGB}{22, 160, 133}
\definecolor{kgColor}{RGB}{41, 128, 185}
\definecolor{tableBg}{RGB}{228, 246, 243}
\definecolor{kgBg}{RGB}{228, 241, 250}
\definecolor{tableDark}{RGB}{14, 102, 85}
\definecolor{kgDark}{RGB}{28, 99, 143}
\title{LLM Inference Enhanced by External Knowledge: A Survey}

\author{
Yu-Hsuan Lin\thanks{Equal contribution.} \quad
Qian-Hui Chen\footnotemark[1] \quad
Yi-Jie Cheng\footnotemark[1] \quad
Jia-Ren Zhang\footnotemark[1] \\
\textbf{Yi-Hung Liu\footnotemark[1] \quad 
Liang-Yu Hsia\footnotemark[1] \quad 
Yun-Nung Chen} \\
National Taiwan University, Taipei, Taiwan \\
\texttt{\{r13922159, r13921039, b09202004, r12921122\}@ntu.edu.tw}\\
\texttt{\{b11611049, r13942129\}@ntu.edu.tw} \quad \texttt{y.v.chen@ieee.org}
}

\begin{document}
\maketitle
\input{contents/abstract}
\input{contents/chapter1}
\input{contents/chapter2}
\input{contents/chapter3}
\input{contents/chapter5}
\input{contents/chapter6}
\input{contents/chapter7}

\input{contents/limitations}
\bibliography{anthology,custom}

\appendix
\clearpage
\input{contents/appendix_benchmarks}

\end{document}

%% file: contents/abstract.tex
\begin{abstract}
Recent advancements in large language models (LLMs) have enhanced natural-language reasoning. However, their limited parametric memory and susceptibility to hallucination present persistent challenges for tasks requiring accurate, context-based inference. To overcome these limitations, an increasing number of studies have proposed leveraging external knowledge to enhance LLMs. This study offers a systematic exploration of strategies for using external knowledge to enhance LLMs, beginning with a taxonomy that categorizes external knowledge into unstructured and structured data. We then focus on structured knowledge, presenting distinct taxonomies for tables and knowledge graphs (KGs), detailing their integration paradigms with LLMs, and reviewing representative methods. Our comparative analysis further highlights the trade-offs among interpretability, scalability, and performance, providing insights for developing trustworthy and generalizable knowledge-enhanced LLMs.\footnote{\url{https://github.com/MiuLab/KG-Survey}}

\end{abstract}

%% file: contents/chapter1.tex
\section{Introduction}
\label{Introduction}

Advancements in LLMs~\cite{GPT2, GPT3, GPT4, Llama, Llama2, Llama3} have significantly advanced natural language processing. These models excel in natural language understanding, generation, and reasoning. However, they face several limitations, prompting the integration of external knowledge to improve performance, reliability, and interpretability.

Key limitations include: 1) outdated knowledge due to the cutoff of training data, 2) hallucinations \cite{Hallucination_survey}, 3) lack of domain-specific expertise, and 4) lack of transparency in responses. To address these, researchers are transforming LLMs into knowledge-aware assistants by incorporating external information during inference.

Despite rapid development, the landscape of knowledge-enhanced LLM inference remains fragmented, with methodologies evolving across various data modalities and domains. Most of these methods follow the retrieve-augment-generate (RAG) \cite{RAG_survey} framework, which focuses on retrieving relevant external knowledge, augmenting it with the model's internal understanding, and generating a response. This approach has become dominant due to its effectiveness in reducing hallucinations while preserving the generative capabilities of LLMs.

In this paper, we start by introducing a taxonomy of knowledge sources, categorized into structured \cite{Table-Survey, KG-Survey, otherstructure-1} and unstructured \cite{gao2024retrievalaugmentedgenerationlargelanguage, Yin_2024, Selmy2024} forms. With a primary focus on structured data due to its explicit relationships and reasoning capabilities, we review methods that take advantage of the data from tables and KGs. This includes symbolic reasoning \cite{Text-to-SQL, NormTab}, neural reasoning \cite{Chain-of-Table}, and hybrid reasoning \cite{TabSQLify, ALTER, Plan-of-SQLs} for tables, and loose coupling \cite{KAPING, CoK, Retrieve-Rewrite-Answer} and tight coupling \cite{ToG, ToG-2, PoG} for KGs. We conclude by comparing these approaches in terms of their strengths, limitations, and key trade-offs.

By focusing on knowledge integration, we emphasize methods that enhance LLMs during inference rather than pre-training or fine-tuning. This approach requires fewer resources and allows for dynamic responses, as external knowledge can be added on-demand without retraining while maintaining model performance.

The main contributions of this survey are:
\begin{compactitem}
\item A comprehensive categorization of external knowledge sources and integration strategies, with a particular focus on tables and knowledge graphs (KGs).
\item A comparative analysis of representative approaches through benchmark experiments, highlighting their strengths, limitations, and trade-offs.
\item Practical insights and guidance to inform future research in knowledge-enhanced LLMs.
\end{compactitem}

%% file: contents/chapter2.tex
\section{A Taxonomy of External Knowledge Sources}
\label{Section2}
\input{contents/chapter2_fig1_new}
We begin by introducing knowledge sources that can be integrated into various approaches. These sources are categorized into structured and unstructured data, as illustrated in \autoref{fig:taxonomy_table_llm}.

\subsection{Unstructured Data} 
\label{subsec:2_unstructured}
Unstructured data represents the most abundant and widely available form of external information, encompassing diverse sources such as free-form text documents, images, audio recordings, and videos \cite{gao2024retrievalaugmentedgenerationlargelanguage, Yin_2024}. Unlike structured data, unstructured data does not conform to any predefined schema or strict relational framework, resulting in high volume, variety, and complexity levels. Consequently, extracting meaningful insights from unstructured sources typically requires advanced interpretation techniques, including NLP, computer vision, speech recognition, and multimodal fusion methods \cite{Selmy2024}. While LLMs are pre-trained on much unstructured text, integrating targeted unstructured data during inference allows them to access information beyond their parametric knowledge, including more timely or specific details.

However, relying on unstructured data also presents notable drawbacks compared to structured data sources. Due to its inherent lack of schema and explicit relational organization, unstructured data often requires computationally expensive preprocessing and sophisticated retrieval techniques to identify relevant information accurately. Additionally, reasoning from raw, free-form text or multimodal data tends to introduce ambiguity and noise, increasing the risk of model hallucinations or inaccurate inferences.

\subsection{Structured Data}
\label{subsec:2_structured}
Structured data refers to information that adheres to a predefined schema, providing an organized representation of facts and relationships crucial for precise reasoning and factual consistency. Compared to unstructured data, structured data offers several advantages, including clearer semantic definitions, more precise information retrieval, and enhanced interpretability.  However, structured data also presents challenges. Integrating structured information with LLMs typically requires transforming data from its structured format into textual or symbolic representations, a process that can lead to potential loss or misinterpretation of information. Moreover, effective interaction with structured data often requires LLMs to integrate external tools, such as SQL engines, to perform complex operations like querying, aggregation, or filtering. Structured data also tends to have high dimensionality and large-scale formats, which can exceed the input length limit of typical LLMs and introduce additional computational challenges. Finally, reasoning over structured data often involves multi-step logical inference or numerical computations, posing difficulties for LLMs that are primarily trained on natural language patterns.

Currently, the most widely explored structured data sources in knowledge-enhanced LLM research are tables and KGs. In this subsection, we will focus on these two prominent structured data sources that are commonly used to enhance LLM inference.

\subsubsection {Tables} 
\label{subsec:2_1_tables}
    Tables~\cite{Table-Survey, Table-Survey-2} are one of the most extensively studied structured data sources for augmenting LLM inference. Their format, organized into rows and columns under semantically meaningful headers, offers a compact yet informative way to represent structured facts. This makes tables especially useful for tasks that require factual verification, numerical reasoning, and structured query understanding~\cite{chen2019tabfact, pasupat2015compositional, nan2022fetaqa}. This straightforward yet highly informative representation facilitates precise data retrieval and supports transparent reasoning processes.

    However, tables also come with inherent limitations. Their rigid format might lack the flexibility to capture more complex or multi-faceted relationships that extend beyond simple row-column structures. Furthermore, accurately converting tabular data into text-based inputs comprehensible to LLMs may sometimes prove challenging, potentially introducing ambiguity or misalignment.
    
    Recent methods that leverage tables to enhance LLM inference include Chain-of-Table~\cite{Chain-of-Table}, TabSQLify~\cite{TabSQLify}, and H-STAR~\cite{H-STAR}. These specific integration strategies and their effectiveness will be discussed in detail in ~\autoref{CH3: Methods for Integrating External Knowledge into LLMs}.

\subsubsection {Knowledge Graphs (KGs)}
    \label{subsec:2_2_kgs}
    KGs~\cite{KG-Survey} represent another significant and widely adopted form of structured data utilized to enhance LLM inference. A KG encodes information as a structured network of interconnected entities. Relationships between these entities are explicitly defined using triples, consisting of a subject, predicate, and object. This structured representation allows KGs to effectively capture relationships between contexts, enabling sophisticated reasoning processes such as multi-hop inference, entity linking, and fact verification tasks~\cite{yih2016value, talmor2018web} that require explicit relational understanding.

    Nonetheless, integrating knowledge graphs with LLMs presents unique challenges. Due to the inherently graph-based nature of KGs, converting their structured relational data into neural representations suitable for LLM is often nontrivial and may lead to potential information loss. Additionally, reasoning directly over large-scale graphs might incur significant computational overhead and complexity.
    
    Recent notable methods exploring KG integration into LLM inference include Chain-of-Knowledge~\cite{CoK}, ToG~\cite{ToG, ToG-2}, and Plan-on-Graph~\cite{PoG}. These methods vary in their integration strategies. Detailed discussions of their specific methodologies will be further detailed in ~\autoref{CH3: Methods for Integrating External Knowledge into LLMs}.

%% file: contents/chapter2_fig1_new.tex
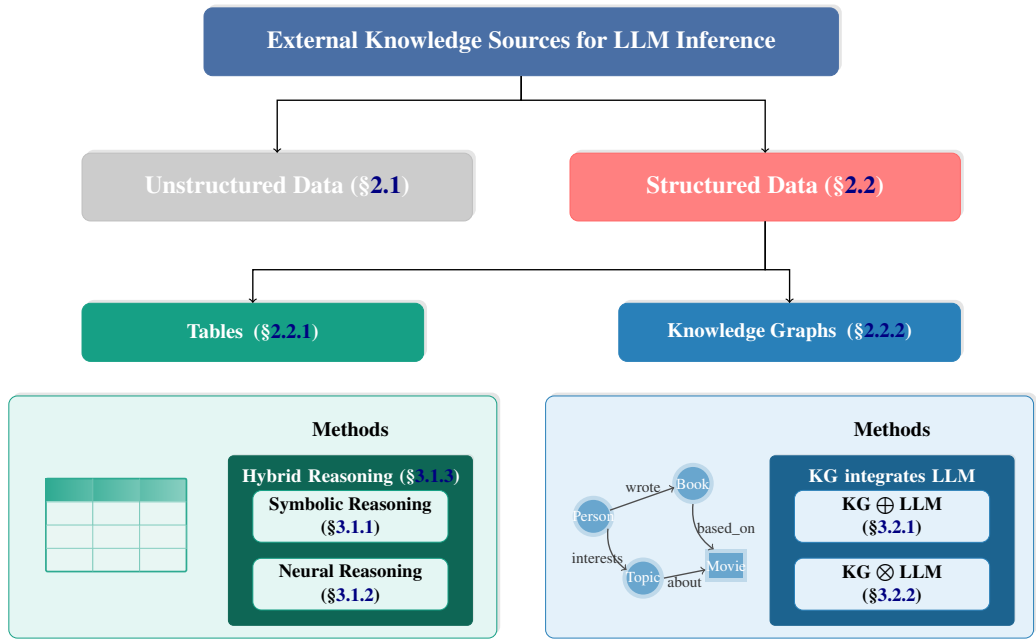
\begin{figure*}[ht]
\centering
\resizebox{0.9\textwidth}{!}{
\begin{tikzpicture}[
    title/.style={
        draw=headerColor, 
        fill=headerColor, 
        rounded corners=5pt, 
        text=white, 
        minimum width=13cm, 
        minimum height=1.4cm, 
        font=\bfseries\Large,
        drop shadow={shadow xshift=2pt, shadow yshift=2pt, opacity=0.2}
    },
    unstructuredsection/.style={
        draw=gray!50, 
        fill=gray!40, 
        rounded corners=5pt, 
        text=white, 
        minimum width=8cm, 
        minimum height=1.4cm, 
        font=\bfseries\Large,
        drop shadow={shadow xshift=2pt, shadow yshift=2pt, opacity=0.2}
    },
    structuredsection/.style={
        draw=red!60, 
        fill=red!50, 
        rounded corners=5pt, 
        text=white, 
        minimum width=8cm, 
        minimum height=1.4cm, 
        font=\bfseries\Large,
        drop shadow={shadow xshift=2pt, shadow yshift=2pt, opacity=0.2}
    },
    tablesection/.style={
        draw=tableColor, 
        fill=tableColor, 
        rounded corners=5pt, 
        text=white, 
        minimum width=7cm, 
        minimum height=1.2cm, 
        font=\bfseries\large,
        drop shadow={shadow xshift=2pt, shadow yshift=2pt, opacity=0.2}
    },
    kgsection/.style={
        draw=kgColor, 
        fill=kgColor, 
        rounded corners=5pt, 
        text=white, 
        minimum width=7cm, 
        minimum height=1.2cm, 
        font=\bfseries\large,
        drop shadow={shadow xshift=2pt, shadow yshift=2pt, opacity=0.2}
    },
    tablecontent/.style={
        draw=tableColor, 
        fill=tableBg, 
        rounded corners=5pt, 
        text=black, 
        minimum width=10cm,
        minimum height=5cm, 
        align=left,
        drop shadow={shadow xshift=2pt, shadow yshift=2pt, opacity=0.2}
    },
    kgcontent/.style={
        draw=kgColor, 
        fill=kgBg, 
        rounded corners=5pt, 
        text=black, 
        minimum width=10cm, 
        minimum height=5cm, 
        align=left,
        drop shadow={shadow xshift=2pt, shadow yshift=2pt, opacity=0.2}
    },
    unstructuredcontent/.style={
        draw=gray!50, 
        fill=gray!10, 
        rounded corners=5pt, 
        text=black, 
        minimum width=8cm, 
        minimum height=3cm, 
        align=left,
        drop shadow={shadow xshift=2pt, shadow yshift=2pt, opacity=0.2}
    },
    tablemethod/.style={
        draw=tableDark, 
        fill=tableDark, 
        rounded corners=3pt, 
        text=white,  
        font=\bfseries,
        drop shadow={shadow xshift=1pt, shadow yshift=1pt, opacity=0.15}
    },
    tablemethodcontent/.style={
        draw=tableColor, 
        fill=tableBg, 
        rounded corners=5pt, 
        text=black,  
        font=\bfseries,
    },
    kgmethod/.style={
        draw=kgDark, 
        fill=kgDark, 
        rounded corners=3pt, 
        text=white, 
        minimum height=1cm, 
        font=\bfseries,
        drop shadow={shadow xshift=1pt, shadow yshift=1pt, opacity=0.15}
    },
    kgmethodcontent/.style={
        draw=kgColor, 
        fill=kgBg, 
        rounded corners=5pt, 
        text=black,  
        font=\bfseries,
    },
    conn/.style={
        thick, 
        ->
    },
    subtitle/.style={
        font=\bfseries, 
        align=center
    }
]

\node[title] (title) at (2,0) {External Knowledge Sources for LLM Inference};

\node[unstructuredsection] (unstructured) at (-3,-3) {Unstructured Data (\S \ref{subsec:2_unstructured})};
\node[structuredsection] (structured) at (7,-3) {Structured Data (\S \ref{subsec:2_structured})};

\draw[conn] (title.south) -- ++(0,-0.5) -| (unstructured.north);
\draw[conn] (title.south) -- ++(0,-0.5) -| (structured.north);

\node[tablesection] (tables) at (-3.5,-6) {Tables~ (\S  \ref{subsec:2_1_tables})};
\node[kgsection] (kg) at (7.5,-6) {Knowledge Graphs~ (\S  \ref{subsec:2_2_kgs})};

\draw[conn] (structured.south) -- ++(0,-1) -| (tables.north);
\draw[conn] (structured.south) -- ++(0,-1) -| (kg.north);

\node[tablecontent, below=0.7cm of tables] (tableContent) {};

\node[subtitle] (tableMethodTitle) at (-1.5,-8) {\large Methods};

\node[tablemethod, minimum width=5cm, minimum height=3.5cm] (method1) at (-1.5,-10.3) {};

\node[align=center, text width=8cm] at (-1.5,-9) {
    \textbf{\textcolor{white}{Hybrid Reasoning (\S \ref{subsec:hybrid})}}
};

\node[tablemethodcontent, align=center, minimum width=4cm] at (-1.5,-9.8) {Symbolic Reasoning\\(\S \ref{subsec:symbolic})};

\node[tablemethodcontent, align=center, minimum width=4cm] at (-1.5,-11.2) {Neural Reasoning\\(\S \ref{subsec:neural})};

\begin{scope}[shift={(-6.3,-9.5)}, scale=1.2]

    \shade[left color=tableColor!90, right color=tableColor!70] 
      (-1.2,0) rectangle (-0.4,0.4);
    \shade[left color=tableColor!80, right color=tableColor!60] 
      (-0.4,0) rectangle (0.4,0.4);
    \shade[left color=tableColor!70, right color=tableColor!50] 
      (0.4,0) rectangle (1.2,0.4);

    \fill[tableColor!10] (-1.2,-0.4) rectangle (-0.4,0);
    \fill[tableColor!10] (-0.4,-0.4) rectangle (0.4,0);
    \fill[tableColor!10] (0.4,-0.4) rectangle (1.2,0);
    
    \fill[white] (-1.2,-0.8) rectangle (-0.4,-0.4);
    \fill[white] (-0.4,-0.8) rectangle (0.4,-0.4);
    \fill[white] (0.4,-0.8) rectangle (1.2,-0.4);
    
    \fill[tableColor!10] (-1.2,-1.2) rectangle (-0.4,-0.8);
    \fill[tableColor!10] (-0.4,-1.2) rectangle (0.4,-0.8);
    \fill[tableColor!10] (0.4,-1.2) rectangle (1.2,-0.8);

    \draw[tableColor!90, line width=1pt, rounded corners=1pt] 
      (-1.2,0.4) rectangle (1.2,-1.2);

    \draw[tableColor!60, line width=0.5pt] (-1.2,0) -- (1.2,0);
    \draw[tableColor!60, line width=0.5pt] (-1.2,-0.4) -- (1.2,-0.4);
    \draw[tableColor!60, line width=0.5pt] (-1.2,-0.8) -- (1.2,-0.8);
    
    \draw[tableColor!60, line width=0.5pt] (-0.4,0.4) -- (-0.4,-1.2);
    \draw[tableColor!60, line width=0.5pt] (0.4,0.4) -- (0.4,-1.2);

    \node at (-0.8,0.2) {\textcolor{white}{\footnotesize}};
    \node at (0,0.2) {\textcolor{white}{\footnotesize}};
    \node at (0.8,0.2) {\textcolor{white}{\footnotesize}};
    
    \node at (-0.8,-0.2) {\textcolor{black!70}{\footnotesize}};
    \node at (0,-0.2) {\textcolor{black!70}{\footnotesize}};
    \node at (0.8,-0.2) {\textcolor{black!70}{\footnotesize}};

    \node at (-0.8,-0.6) {\textcolor{black!70}{\footnotesize }};
    \node at (0,-0.6) {\textcolor{black!70}{\footnotesize }};
    \node at (0.8,-0.6) {\textcolor{black!70}{\footnotesize }};
    
    \node at (-0.8,-1.0) {\textcolor{black!70}{\footnotesize }};
    \node at (0,-1.0) {\textcolor{black!70}{\footnotesize }};
    \node at (0.8,-1.0) {\textcolor{black!70}{\footnotesize }};

    \fill[tableColor!30, opacity=0.3] (-1.2,-0.8) rectangle (1.2,-0.4);
\end{scope}

\node[kgcontent, below=0.7cm of kg] (kgContent) {};

\node[subtitle] (kgResearchTitle) at (9.6,-8) {\large Methods};

\node[kgmethod, minimum width=5cm, minimum height=3.5cm] (research2) at (9.6,-10.3) {};
\node[align=center, text width=8cm] at (9.6,-9) {
    \textbf{\textcolor{white}{KG integrates LLM}}
};

\node[kgmethodcontent, align=center, minimum width=4cm] at (9.6,-9.8) {KG \(\bigoplus\) LLM\\(\S \ref{subsec:loose})};

\node[kgmethodcontent, align=center, minimum width=4cm] at (9.6,-11.2) {KG \(\bigotimes\) LLM\\(\S \ref{subsec:tight})};

\begin{scope}[shift={(4.5,-9.8)}, scale=0.85]

    \fill[kgColor!30] (-1.2,0) circle (0.5);
    \fill[kgColor!70] (-1.2,0) circle (0.4);
    
    \fill[kgColor!30] (1.2,0.8) circle (0.5);
    \fill[kgColor!70] (1.2,0.8) circle (0.4);
    
    \fill[kgColor!30] (2.5,-0.8) rectangle (1.5,-1.6);
    \fill[kgColor!70] (2.4,-0.9) rectangle (1.6,-1.5);
    
    \fill[kgColor!30] (0,-1.5) circle (0.5);
    \fill[kgColor!70] (0,-1.5) circle (0.4);
    
    \draw[->, thick, black!70] (-0.8,0.15) -- (0.75,0.7);
    
    \draw[->, thick, black!70] (-0.85,-0.3) to[bend right] (-0.4,-1.3);
    
    \draw[->, thick, black!70] (0.5,-1.5) -- (1.5,-1.3);
    
    \draw[->, thick, black!70] (1.2,0.3) to[bend right] (1.7,-0.8);
    
    \node at (-1.2,0) {\textcolor{white}{\small Person}};
    \node at (1.2,0.8) {\textcolor{white}{\small Book}};
    \node at (2,-1.2) {\textcolor{white}{\small Movie}};
    \node at (0,-1.5) {\textcolor{white}{\small Topic}};
    
    \node at (0,0.7) {\textcolor{black!80}{\footnotesize wrote}};
    \node at (-1.1,-1) {\textcolor{black!80}{\footnotesize interests}};
    \node at (1,-1.6) {\textcolor{black!80}{\footnotesize about}};
    \node at (2,-0.3) {\textcolor{black!80}{\footnotesize based\_on}};
\end{scope}

\end{tikzpicture}
}
\caption{Taxonomy of external knowledge source.}
	\label{fig:taxonomy_table_llm}
\end{figure*}

%% file: contents/chapter3.tex
\section{External Knowledge Integration}
\label{CH3: Methods for Integrating External Knowledge into LLMs}

\input{contents/chapter3_table_fig}

In \autoref{Section2}, we introduced a taxonomy of external knowledge sources, categorizing them according to their structure. We now turn our attention to exploring various methodological strategies for integrating these knowledge sources into LLMs. 

Among the different knowledge sources, structured data, such as Tables and KGs, has gained increasing research interest due to their ability to represent rich, interconnected information. Unlike unstructured data, which lacks clear relational structures, structured data models relationships between entities, enabling more precise reasoning and supporting complex queries \cite{Knowledge_Graph, Table-Survey-2}.

In this section, we focus specifically on Tables and KGs, as they offer a particularly promising avenue for enhancing LLMs through their structured and highly organized representation of knowledge. We will explore Tables and KGs in detail, following the structure provided in \autoref{fig:taxonomy_table_llm}.

\subsection{Table Integration}
\label{sec3-1:table}

Integration methods for tables vary according to their dominant reasoning modality—specifically, whether they rely on \textit{symbolic reasoning}, \textit{neural reasoning}, or a \textit{hybrid combination} of the two. This classification reflects a key design trade-off in table question answering: Whether to reason using predefined rules and logic, such as SQL, or with LLMs. 
A conceptual overview is presented in \autoref{CH3:Table_Fig}, with additional explanations provided in the following sections.

\subsubsection{Symbolic Reasoning} 
\label{subsec:symbolic}
Symbolic reasoning methods rely on LLMs to generate explicit symbolic programs—typically SQL queries, which are then executed over tables to obtain the final answer. For example, Text-to-SQL~\cite{Text-to-SQL} directly translates natural language questions into executable SQL, delegating the reasoning process to the query engine. NormTab~\cite{NormTab} further enhances this pipeline by normalizing web tables before query generation, improving accuracy and robustness under noisy or inconsistent table formats. These symbolic methods offer high precision and strong interpretability. However, symbolic approaches often struggle with questions that require nuanced semantic understanding, common-sense knowledge, or implicit reasoning, where SQL alone is insufficient to fully express the query intent.

\subsubsection{Neural Reasoning} 
\label{subsec:neural}
Neural reasoning methods rely entirely on the LLM’s internal capacity to perform reasoning within the language space, without invoking external symbolic executors. These approaches often follow an end-to-end paradigm, where the model directly predicts the answer conditioned on the question and table input. Typical strategies include few-shot prompting~\cite{Few-Shot} and chain-of-thought (CoT)~\cite{Chain-of-Thought} reasoning. For instance, Chain-of-Table~\cite{Chain-of-Table} enables multi-step table reasoning by prompting the LLM to generate intermediate table operations as text. These neural reasoning methods excel in handling complex, ambiguous, or common-sense reasoning tasks, thanks to their flexible and adaptive nature. However, they also face notable challenges, including hallucination, limited logical reasoning capabilities, and a lack of interpretability.

\subsubsection{Hybrid Reasoning}
\label{subsec:hybrid}
Hybrid reasoning methods combine symbolic execution (e.g., SQL) with neural reasoning, allowing LLMs to handle complex table tasks by leveraging both precise computation and semantic understanding. A common pattern involves the LLM generating SQL to retrieve relevant table subsets, followed by further reasoning over the results, as seen in TabSQLify~\cite{TabSQLify} and ALTER~\cite{ALTER}. Other methods such as Plan-of-SQLs~\cite{Plan-of-SQLs} and Binder~\cite{Binder} use LLMs to produce multi-step symbolic programs or executable code, enabling structured, interpretable reasoning. ReAcTable~\cite{ReAcTable} and DATER~\cite{DATER} adopt iterative pipelines that alternate between LLM-driven reasoning and symbolic tool use, while ProTrix~\cite{ProTrix} and H-STAR~\cite{H-STAR} dynamically choose between neural and symbolic operations at each reasoning step. These hybrid methods leverage the strengths of both symbolic and neural reasoning to overcome the limitations of either approach, making them particularly suitable for complex reasoning.

\subsection{Knowledge Graph (KG) Integration}
\input{contents/chapter3_KG_fig}
\label{subsec:3_2_kg}
    
    KG integration approaches vary based on the tightness of coupling between the LLM and the KG. We classify these methods into loose coupling and tight coupling strategies, reflecting the degree of interaction between LLMs and graph-structured knowledge illustrated in \autoref{CH3:KG_Fig}.

    \subsubsection{Loose Coupling KG \(\bigoplus\) LLM.}
    \label{subsec:loose}
    Loose coupling treats KGs and LLMs as independent components that sequentially interact. 
    Approaches such as CoK \cite{CoK} improve prompting by refining rationales, which generates rationales based on Chain-of-Thought \cite{CoT} and enhances them by retrieving knowledge from KGs.
    Retrieve-Rewrite-Answer (RRA) \cite{Retrieve-Rewrite-Answer} retrieves KG triples and rewrites them into informative natural language prompts for the LLM. Other notable examples include KAPING \cite{KAPING} and DKG \cite{DKG}, both constructing enriched prompts directly from KGs. These methods typically offer simplicity and modularity but may suffer from limited interactive reasoning capability.

    \subsubsection{Tight Coupling KG \(\bigotimes\) LLM.}
    \label{subsec:tight}
    Tight coupling integrates KGs and LLMs more deeply through interactive reasoning loops, enabling dynamic entity and relation exploration within KGs. It treats the LLM as an agent that interactively explores related entities and relations in the KG, using retrieved knowledge for reasoning. This results in more dynamic and accurate retrieval.
    Recently, researchers have increasingly used this approach to integrate LLMs and KGs. Methods such as Think-on-Graph (ToG) \cite{ToG,ToG-2} and StructGPT \cite{StructGPT} actively leverage LLMs to iteratively traverse and reason over graph structures.
    
    Extensions like Plan-on-Graph (PoG) \cite{PoG}, Decoding-on-Graph (DoG) \cite{DoG}, and Generate-on-Graph (GoG) \cite{GoG} further enhance performance by incorporating self-correction, richer context, and complementary internal knowledge, respectively. Despite higher complexity, tight coupling typically yields more accurate and robust reasoning

%% file: contents/chapter3_table_fig.tex
\begin{figure*}[ht]
    \centering
    \includegraphics[width=0.9\textwidth, trim=1.6cm 9.2cm 2.9cm 1.3cm, clip]{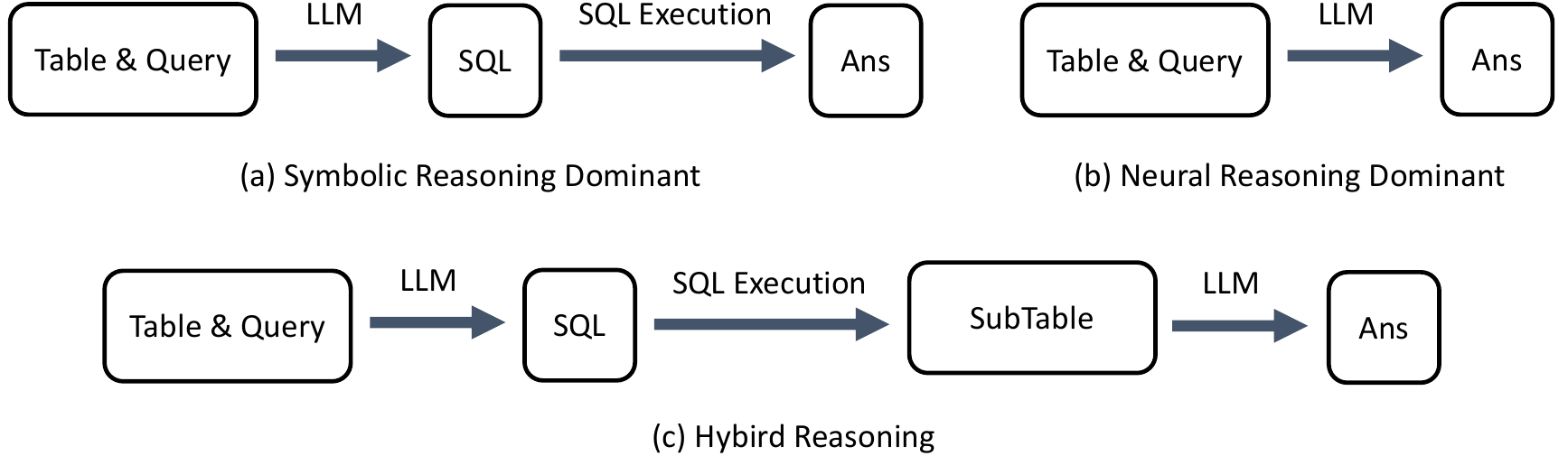}
    \caption{\textbf{Overview of table reasoning paradigms}:
    (a) Symbolic reasoning, where the LLM generates and executes SQL queries over the table. 
    (b) Neural reasoning, where the LLM directly reasons over the table and question in natural language. 
    (c) Hybrid reasoning, where the LLM combines SQL-based subtable extraction with further neural reasoning.}
    \label{CH3:Table_Fig}
\end{figure*}

%% file: contents/chapter3_KG_fig.tex
\begin{figure}
    \centering
    \includegraphics[width=0.52\textwidth, trim=7cm 4.3cm 6cm 4.3cm, clip]{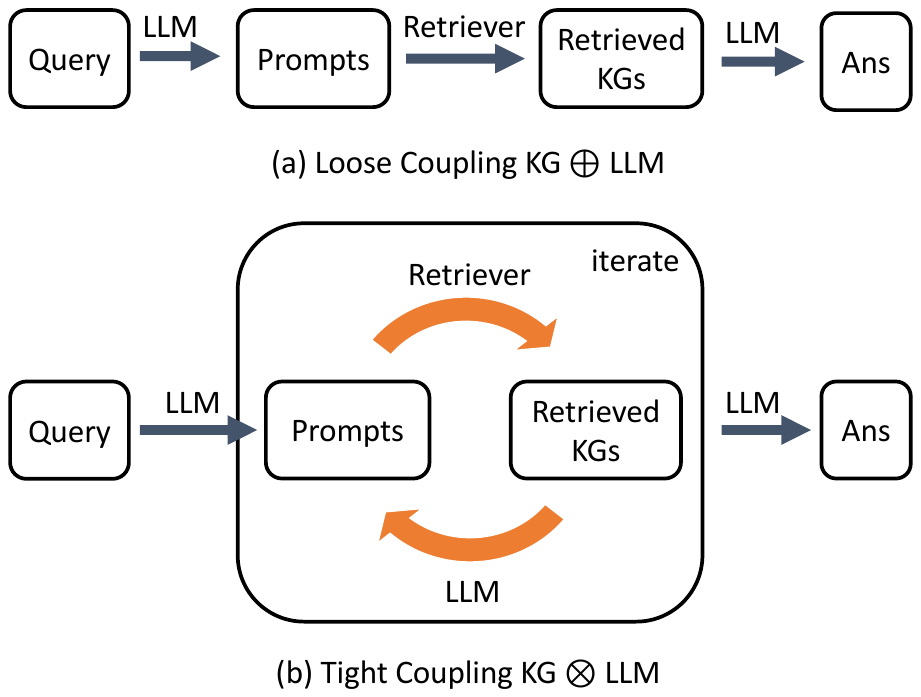}
    \caption{\textbf{Overview of KG-LLM integration strategies}:
    (a) \textit{Loose coupling}: the retrieved KG is directly used by the LLM without refinement.
    (b) \textit{Tight coupling}: the LLM actively refines or integrates retrieved KG information.
    In both strategies, the retriever is often implemented using an LLM-based model.}
    \label{CH3:KG_Fig}
\end{figure}

%% file: contents/chapter5.tex
\section{Discussions}
\label{Discussion}
Motivated by the growing interest in knowledge-enhanced methods, we investigate the trade-offs of recent approaches that leverage structured knowledge resources across diverse scenarios. To ensure a fair comparison, we benchmark the performance of various approaches under a unified experimental setting.
The experimental details can be found in Appendix~\ref{Evaluation Metrics for Knowledge-Augmented LLMs}.

\input{contents/chapter5_experiment_result_table}
\input{contents/chapter5_fig}

\subsection{Table-Enhanced LLM}



To validate the strengths and limitations discussed in \autoref{sec3-1:table}, we reference GPT-3.5-turbo evaluation results reported in prior work on TabFact \cite{chen2019tabfact} and WikiTQ \cite{pasupat2015compositional} datasets. \autoref{table:performance} summarizes the performance across Symbolic, Neural, and Hybrid methods, evaluated using exact match (EM) and binary classification accuracy.
To further contextualize these results, we examine how task-specific characteristics affect performance, in TabFact and WikiTQ respectively.

\paragraph{TabFact} 
Unlike standard QA tasks that focus on answer retrieval, TabFact focuses on fact verification.
However, symbolic methods that rely on SQL generation are inherently more constrained in this setting, as validating the correctness of a statement through SQL is more challenging than retrieving an answer from a table.
As shown in \autoref{table:performance}, neural methods tend to outperform symbolic ones on TabFact, likely due to their strength in modeling semantic nuances and complex reasoning.

\paragraph{WikiTQ} 
To better understand how hybrid methods differ in practice, we compare TabSQLify and H-STAR on the WikiTQ dataset. As shown in \autoref{CH5:Fig}, the question \textit{“How many consecutive seasons premiered in October?”} requires multi-row reasoning over a temporal pattern in the table, which poses challenges for both symbolic and neural methods.

TabSQLify~\cite{TabSQLify} handles this by generating a SQL query to extract October entries and check for consecutive rows. However, as highlighted in the red box, its query fails to capture all necessary rows, leading to an incorrect answer (“1” instead of “4”). This reveals a key limitation: symbolic errors in early stages can mislead the neural component.

H-STAR~\cite{H-STAR}, on the contrary, integrates symbolic and neural signals during extraction, allowing the two to support each other. This design helps preserve important table columns and results in a correct answer.

This example illustrates how hybrid systems vary significantly in their internal design and coordination mechanisms.
Overall, hybrid reasoning currently delivers the most robust performance across diverse question types and datasets, as it combines the precision of symbolic execution with the flexibility of LLM-based neural inference. However, this advantage is not inherent; the effectiveness of a hybrid pipeline ultimately depends on how well its symbolic and neural components are coordinated.



\input{contents/chapter5_kg}

\subsection{Knowledge Graph (KG)-Enhanced LLM}
We examine recent KG-enhanced strategies from two primary perspectives: \textit{performance} and \textit{efficiency}, and analyze the integration methods based on the coupling strategies between KGs and LLMs: loose coupling and tight coupling. A comprehensive overview is provided in \autoref{CH5:KG}.

\subsubsection{Loose Coupling KG \(\bigoplus\) LLM.}

When evaluating performance, KAPING \cite{KAPING} achieves strong zero-shot performance by directly injecting KG facts without retraining, improving accuracy by up to 48\% on benchmarks like WebQSP and Mintaka. RRA \cite{Retrieve-Rewrite-Answer} enhances LLMs by rewriting KG subgraphs into answer-sensitive text, improving knowledge usability. CoK \cite{CoK} goes a step further, promoting factual accuracy across knowledge-intensive tasks by verifying not only final answers but also intermediate reasoning steps.

From an efficiency standpoint, KAPING is lightweight and deployment-friendly. It requires no fine-tuning and performs a single LLM pass, keeping latency low. In contrast, RRA involves multiple LLM calls—retrieval, rewriting, and answering—and may need fine-tuning, introducing additional computational cost. CoK is the most resource-intensive, involving multiple stages such as rationale generation, query rewriting, and answer synthesis, though selective triggering helps mitigate its overhead.

These methods also differ in scope. KAPING is tailored for zero-shot factual QA via direct KG access. RRA focuses on improving KGQA through better knowledge representation. CoK addresses broader goals, including reducing hallucination and ensuring factual consistency across diverse tasks. Unlike KAPING and RRA, which rely on explicit KG retrieval, CoK supports more flexible, general-purpose reasoning, even with incomplete or loosely structured knowledge.

Together, these comparisons highlight a trade-off between simplicity, performance, and generality in KG-enhanced LLMs. Future advances may lie in balancing these dimensions to build systems that are both efficient and broadly reliable.

\noindent
\subsubsection{Tight Coupling (KG \(\bigotimes\) LLM)}

As discussed before, StructGPT \cite{StructGPT} is an early example of tight coupling. It relies on a fixed Iterative Reading-then-Reasoning (IRR) process and pre-defined interfaces. While it improves performance by handling structured inputs, its rigid design limits flexibility, making it less suitable for complex, multi-hop reasoning. ToG \cite{ToG} builds on this idea by treating the LLM as an agent that actively explores the KG, allowing for more dynamic and interpretable reasoning. It also improves efficiency by avoiding additional training. To handle KG incompleteness, ToG-2.0 \cite{ToG-2} introduces a hybrid RAG approach that combines structured KGs with unstructured text. This increases retrieval complexity but enhances reasoning completeness and factual accuracy.

Despite recent advances, issues like fixed path width and irreversible exploration remain.
Plan-on-Graph (PoG) \cite{PoG} addresses these issues by introducing self-correction, enhancing both accuracy and efficiency. 
GoG \cite{GoG} tackles KG incompleteness by incorporating internal LLM knowledge, yielding notable benefits. In contrast, DoG \cite{DoG} employs a well-structured reasoning chain and demonstrates strong performance even with smaller LLMs, effectively leveraging both KG and LLM interpretability.


In summary, DoG is a better fit when resources are limited or smaller models are required. GoG, on the other hand, is more suitable when aiming for optimal accuracy using larger or more advanced LLMs, particularly in real-world applications.


%% file: contents/chapter5_experiment_result_table.tex
\begin{table}[t!]
    \centering
    \small
    \begin{tabular}{lcccc}
    \toprule
     & \textbf{WikiTQ} & \textbf{TabFact} \\
    \midrule
    \multicolumn{3}{l}{\it Symbolic Reasoning}     \\
    Text-to-SQL & 51.3 & 62.3 \\
    NormTab & 61.2 & 68.9  \\

    \midrule
    \multicolumn{3}{l}{\it{Neural Reasoning}} \\
    End-to-End QA$^{\dagger}$ & 51.8 & 70.5 \\
    Few-shot QA$^{\dagger}$ & 52.6 & 71.5 \\
    CoT$^{\dagger}$ & 53.5 & 65.4 \\
    Chain-of-Table & 59.9 & 80.2  \\
    
    \midrule
    \multicolumn{3}{l}{\it{Hybrid Reasoning}} \\
    BINDER$^{*}$ & 55.4 & 79.1  \\
    ReAcTable & 52.4 & 73.1 \\
    DATER$^{*}$ & 52.8 & 78.0  \\
    TabSQLify & 64.7 & 79.5  \\
    ALTER &  67.4 &  84.3 \\
    Plan-of-SQLs & 54.8 & 78.3 \\
    ProTrix & 65.2 & 83.5 \\
    H-STAR & \textbf{69.6} & \textbf{85.0} \\
    
    \bottomrule
    \end{tabular}
\caption{GPT-3.5-Turbo results on question answering and fact verification tasks. Bold indicates best performance. $^{*}$: results reported by \citet{TabSQLify} with self-consistency by running multiple inference iterations, 20 for DATER on both datasets, and 50 for BINDER on TabFact. $^{\dagger}$: results reported by \citet{H-STAR}; Text-to-SQL results reported by \citet{NormTab}.} 
\label{table:performance}
\end{table}


%% file: contents/chapter5_fig.tex
\begin{figure*}[htbp]
    \centering
    \includegraphics[width=\textwidth]{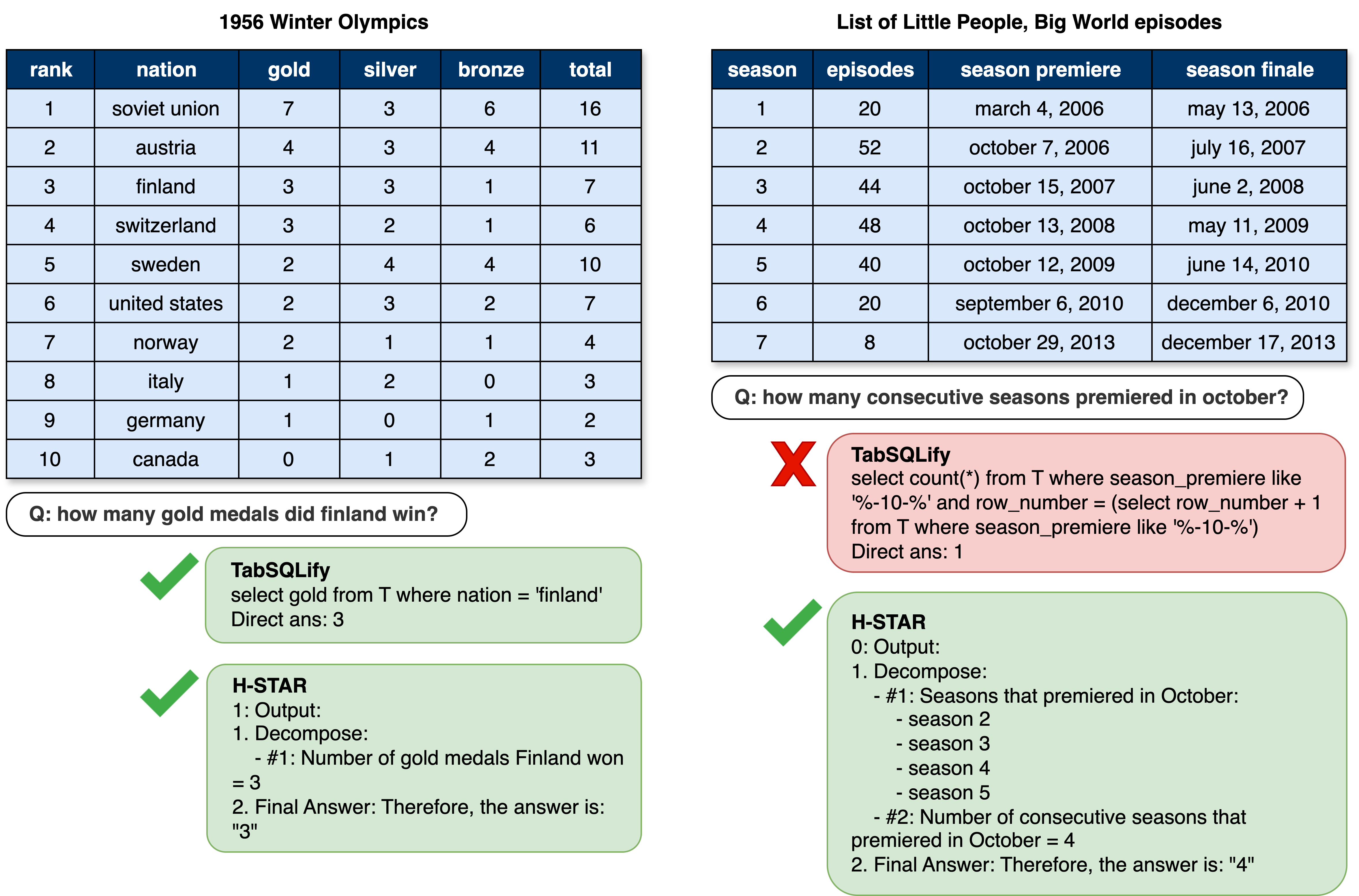}
    \caption{Illustration of two example questions from WikiTQ. The left example can be directly answered via SQL, while the right example requires multi-step reasoning over temporal information. TabSQLify succeeds only in the left case, whereas H-STAR answers both questions correctly.}
    \label{CH5:Fig}
\end{figure*}

%% file: contents/chapter5_kg.tex
\begin{table*}[ht]
\centering
\small
\resizebox{\textwidth}{!}{
    \begin{tabular}{lp{6cm}p{4cm}p{4cm}}
    \toprule
    \bf Methods     & \bf Objective & \bf Performance & \bf Efficiency\\ 
    \midrule
    \multicolumn{4}{l}{\textit{Loose Coupling (KG \(\bigoplus\) LLM).}} \\
    CoK         & Enhance factual accuracy by integrating diverse knowledge sources.
                & Improves factual consistency; no SOTA claims.
                & Moderate; involves multi-stage processing.
                \\
    KAPING      & Zero-shot KGQA without model training.
                & Achieves up to 48\% improvement over baselines; SOTA in zero-shot settings.
                & High; simple retrieval and prompt augmentation.
                \\
    RRA         & Transform KG data into text for LLM-based QA.
                & Shows performance improvements; no SOTA claims.
                & High; straightforward KG-to-text conversion.
                \\ 
    \midrule
    \multicolumn{4}{l}{\textit{Tight Coupling (KG \(\bigotimes\) LLM)}}\\
    ToG         & Deep reasoning over KGs using LLMs as agents.
                & Achieves SOTA in multiple datasets without additional training.
                & Moderate; involves beam search over KGs.
                \\
    ToG-2.0     & Integrate structured and unstructured knowledge for deep reasoning.
                & Achieves SOTA on 6 out of 7 datasets; enhances smaller models to match larger ones.
                & Moderate; iterative retrieval from multiple knowledge sources.
                \\
    StructGPT   & Zero-shot reasoning over structured data.
                & Enhances reasoning capabilities; no SOTA claims.
                & Moderate; iterative reading and reasoning process.
                \\
    PoG         & Focus on self-correction and the adaptive breadth. 
                & Better than ToG.              
                & Lower; involves multiple iterations for self-correction.
                \\
    GoG         & Handle incomplete KG scenarios by generating missing knowledge.
                & Better for larger models. 
                & Not mentioned.
                \\
    DoG         & Aims to create a well-formed chain to enhance the integration of LLM reasoning and KG structure. 
                & Better for smaller models.
                & Not mentioned.
                \\ 
    \bottomrule
    \end{tabular}
}
\caption{Comparison among different KG-based methods.}
\label{CH5:KG}
\end{table*}

%% file: contents/chapter6.tex
\section{Challenges and Future Directions}

\paragraph{Error Propagation}
As discussed in the context of table-enhanced LLMs, hybrid reasoning methods have emerged as a promising direction due to their ability to combine the complementary strengths of symbolic and neural approaches. However, our analysis reveals a critical issue about error propagation. If the symbolic module produces incorrect intermediate results, the subsequent neural reasoning may be misled, potentially causing the overall reasoning process to fail. A key future direction is to develop more dynamic and fault-tolerant hybrid architectures that allow flexible collaboration between symbolic and neural components without rigid dependencies.


\paragraph{Balancing Input Size and Information Loss}
Another practical challenge concerns the input size limitations of LLMs, especially when processing large tables. Many existing methods employ table extraction techniques to filter out irrelevant cells and reduce input size. While often effective, such filtering risks excluding critical information, which can degrade performance. Therefore, a central open problem is how to balance input reduction with the risk of over-pruning, preserving essential content without exceeding model constraints.

\paragraph{Efficiency and System Complexity}
Beyond accuracy and input constraints, computational efficiency remains a crucial consideration. Although hybrid approaches often outperform purely symbolic or neural methods by leveraging their respective strengths, this integration typically comes with increased system complexity and latency. Whether the symbolic and neural modules are executed sequentially or in parallel, the resulting hybrid system tends to be more resource-intensive. Future research should explore how to preserve the benefits of both paradigms while minimizing computational overhead, potentially through more selective, adaptive, or lightweight coordination mechanisms.

\paragraph{Multimodal Knowledge}
Integrating multimodal structured data represents a pivotal yet underexplored frontier for KG-enhanced LLMs. While current pipelines can incorporate textual triples and numerical tables, they seldom support the systematic integration of additional modalities such as images, audio, or video metadata. To advance this direction, future work should focus on developing robust alignment protocols that accurately map heterogeneous signals to graph entities with minimal semantic loss. Such cross-modal alignment will be crucial for enabling effective reasoning over shared, multimodal representations.



\paragraph{Real-Time Reasoning}
Real-time reasoning poses another critical challenge, especially in dynamic domains like finance, healthcare, and epidemiology, where knowledge evolves rapidly. Static knowledge snapshots quickly become outdated, limiting their practical utility. Future systems must incorporate stream-level retrieval of incremental updates, paired with low-latency adaptation strategies, such as lightweight parameter tuning or dynamic prompting, to ensure outputs remain temporally consistent. Additionally, implementing rigorous version control and uncertainty calibration mechanisms will be essential to maintain trustworthy, time-sensitive inference.


%% file: contents/chapter7.tex
\section{Conclusion}
\label{Conclusion}
In this survey, we explored the integration of external knowledge into LLM inference, focusing on structured knowledge in the form of tables and KGs. We presented a taxonomy to highlight their key characteristics. We also conducted an in-depth analysis of representative methods, identifying key challenges and outlining promising directions for future research.

%% file: contents/limitations.tex
\section*{Limitations}

Despite our effort to categorize and analyze tables and KGs from our perspective, our discussion has several limitations. 

First, when comparing the performance of table-based approaches, we did not account for the impact of input table size. Since table size can significantly influence model performance—affecting both attention efficiency and reasoning accuracy, future work should systematically investigate how varying table scales affect integration strategies and reasoning effectiveness.

Second, we did not explicitly address the challenge of multi-hop complexity, which can significantly increase reasoning difficulty and computational cost in KG-augmented LLMs.

Lastly, we did not conduct a direct comparative analysis between table-based and KG-based reasoning methods. These two forms of structured knowledge differ significantly in format, access granularity, and reasoning patterns. A more unified evaluation framework would be valuable for understanding their respective advantages and for guiding future hybrid designs that incorporate both.

%% file: contents/appendix_benchmarks.tex
\section{Benchmarks and Evaluation Metrics of Knowledge-Augmented LLMs}
\label{Evaluation Metrics for Knowledge-Augmented LLMs}

To systematically evaluate the reasoning capabilities of LLMs augmented with external knowledge, it is important to understand how they are assessed across diverse tasks and modalities. This section provides an overview of widely used benchmark datasets and evaluation metrics for measuring the effectiveness of knowledge-enhanced LLM inference.

We categorize the benchmarks into three major task types, corresponding to the knowledge source utilized. \autoref{tab:evaluation-metrics} summarizes representative benchmarks along with the associated evaluation metrics for each task category, offering a structured view of how knowledge-augmented LLMs are assessed under different reasoning scenarios.

\begin{table*}[t!]
\centering
\small
\begin{tabular}{lp{5cm} p{6.4cm}}
\toprule
\textbf{Task Type} & \textbf{Benchmarks} & \textbf{Common Evaluation Metrics} \\
\midrule
\textbf{General Knowledge QA} & 
TriviaQA, HotpotQA, FreebaseQA & Exact Match (EM), F1, Recall, BERTScore, GPT-4 Average Ranking  \\
\midrule
\textbf{Table-Based Reasoning} & 
TabFact, WikiTQ, FeTaQA, WikiSQL & 
Binary Classification Accuracy, Exact Match (EM), ROUGE \\
\midrule
\textbf{Graph-Based Reasoning} & 
WebQSP, CWQ, MetaQA & 
Hits@1, Hits@K, Path Accuracy, Entity Linking Accuracy \\
\bottomrule
\end{tabular}
\caption{Benchmark datasets and evaluation metrics for knowledge-augmented LLMs across various task types.}
\label{tab:evaluation-metrics}
\end{table*}

\subsection{General Knowledge Tasks}
Open-domain QA tasks evaluate a model’s ability to extract and reason over unstructured textual information, typically from sources such as Wikipedia or the open web. 
\begin{compactitem}
\item TriviaQA \cite{joshi2017triviaqa} includes trivia-style questions paired with evidence documents and primarily tests the model’s ability to retrieve and synthesize a single relevant fact. 
\item HotpotQA \cite{yang2018hotpotqa} increases task complexity by introducing multi-hop reasoning, requiring the model to aggregate information from multiple supporting documents, making it a valuable benchmark for testing whether a model can synthesize and reason across separate sources.
\item FreebaseQA \cite{jiang-etal-2019-freebaseqa} comprises factoid questions whose answers are grounded in the Freebase knowledge graph. However, since it does not provide structured queries or reasoning paths, it is typically used as an open-domain QA dataset rather than a KGQA benchmark.
\end{compactitem}

Common evaluation metrics include:
\begin{compactitem}
\item Exact Match (EM): strict string equality between predicted and ground truth answers.
\item F1 score: token-level overlap between predictions and references. 
\item Recall: particularly for retrieval coverage.
\item BERTScore: measuring semantic similarity using contextual embeddings.
\item GPT-4 Average Ranking: a preference-based metric reflecting model output quality as judged by GPT-4. 
\end{compactitem}

\subsection{Table-Based Reasoning}
\label{subsec:4_2_table}
Table reasoning tasks evaluate a model's ability to understand, retrieve, and reason over table-structured data. 
\begin{compactitem}
\item WikiTableQuestions (WikiTQ) \cite{pasupat2015compositional} is a large-scale benchmark designed to evaluate models on question answering over semi-structured tables from Wikipedia, requiring capabilities such as comparison, aggregation, and arithmetic reasoning. 
\item TabFact \cite{chen2019tabfact} is a benchmark that evaluates a model's ability to determine whether a natural language statement is entailed or refuted by tables from Wikipedia, requiring linguistic understanding and symbolic reasoning. 
\item FeTaQA \cite{nan2022fetaqa} focuses on complex reasoning and generating explanations using Wikipedia tables. 
\item WikiSQL \cite{zhong2017seq2sql} enables natural language interfaces to relational databases by translating questions into SQL queries, allowing users to access data without explicit knowledge of SQL.
\end{compactitem}

Common evaluation metrics include:
\begin{compactitem}
\item Binary Classification Accuracy: for fact verification tasks
\item Exact Match (EM): applied in table-based QA for precise string matching. 
\item ROUGE: measures n-gram overlap between generated and reference textual explanations.
\end{compactitem}


\subsection{Graph-Based Reasoning}
\label{subsec:4_3_kg}
Graph-based reasoning benchmarks assess a model’s ability to perform structured inference using KGs. These tasks emphasize multi-hop, entity-centric, and relation-aware reasoning.
\begin{compactitem}
\item WebQSP \cite{yih2016value} is a widely used benchmark for semantic parsing into SPARQL queries, testing both single-hop and multi-hop reasoning over entity relationships. It evaluates models for converting complex questions into executable formal logic.
\item ComplexWebQuestions (CWQ) \cite{talmor2018web} extends WebQuestions \cite{berant2013semantic} with compositional and conjunctive questions requiring integration of multiple KG facts from web snippets. It tests models on combining KG reasoning with retrieval-based evidence, making it valuable for evaluating hybrid reasoning that bridges symbolic and textual knowledge.
\item MetaQA \cite{puerto-etal-2023-metaqa} tests multi-hop KGQA using movie-related triples, covering 1-hop, 2-hop, and 3-hop queries.
\end{compactitem}

Common evaluation metrics include:
\begin{compactitem}
\item Hits@1, Hits@K: Common evaluation metrics include Hits@1 and Hits@K for assessing the proportion of correct entities ranked within the top positions.
\item Path Accuracy: for evaluating the correctness of inferred reasoning paths.
\item Entity Linking Accuracy: for measuring the precision of entity disambiguation.

\end{compactitem}